\documentclass{article}

\usepackage[final,nonatbib]{nips_2017}

\usepackage[final,nonatbib]{nips_2017}

\usepackage{todonotes}
\usepackage[utf8]{inputenc} 
\usepackage[T1]{fontenc}    
\usepackage{hyperref}       
\usepackage{url}            
\usepackage{booktabs}       
\usepackage{amsfonts}       
\usepackage{nicefrac}       
\usepackage{microtype}      
\usepackage{tabularx, booktabs,adjustbox}
\usepackage{multirow}
\usepackage{caption}
\usepackage{subcaption}
\usepackage{array}
\usepackage[affil-it]{authblk}
\usepackage{float}
\usepackage{multirow}
\usepackage{footnote}
\usepackage[super,sort&compress,comma]{natbib}
\usepackage{decimal}
\usepackage{framed}

\usepackage{lineno}

\hypersetup{
    colorlinks,
    citecolor=black,
    linkcolor=magenta,
}

\newcolumntype{Y}{>{\centering\arraybackslash}X}

\begin{document}

\title{%
   \Large Can Artificial Intelligence Reliably Report Chest X-Rays? \\
   \large \normalfont{Radiologist Validation of an Algorithm trained on 2.3 Million X-Rays}}

\author[1]{Preetham Putha}
\author[1]{Manoj Tadepalli}
\author[1]{Bhargava Reddy}
\author[1]{Tarun Raj}
\author[1]{Justy Antony Chiramal}
\author[2]{Shalini Govil}
\author[2]{Namita Sinha}
\author[2]{Manjunath KS}
\author[2]{Sundeep Reddivari}
\author[1]{Ammar Jagirdar}
\author[1]{Pooja Rao}
\author[1]{Prashant Warier}

\affil[1]{Qure.ai, Mumbai, IN}
\affil[2]{Columbia Asia Radiology Group, Bangalore, IN}

\newcommand{\absdiv}[1]{%
  \par\addvspace{0.5\baselineskip}
  \noindent\textbf{#1}\quad\ignorespaces
}

\maketitle
\begin{abstract}
\small
\absdiv{Background and Objectives}
Chest X-rays are the most commonly performed, cost-effective diagnostic imaging tests ordered by physicians. A clinically validated, automated artificial intelligence system that can reliably separate normal from abnormal would be invaluable in addressing the problem of reporting backlogs and the lack of radiologists in low-resource settings. The aim of this study was to develop and validate a deep learning system to detect chest X-ray abnormalities.

\absdiv{Methods}
A deep learning system was trained on 2.3 million chest X-rays and their corresponding radiology reports to identify abnormal X-rays and the following specific abnormalities: blunted costophrenic angle, cardiomegaly, cavity, consolidation, fibrosis, hilar enlargement, nodule, opacity and pleural effusion. The system was tested against - 1. A three-radiologist majority on an independent, retrospectively collected set of 2000 X-rays(CQ2000) 2. The radiologist report on a separate validation set of 100,000 scans(CQ100k). The primary accuracy measure was area under the ROC curve (AUC), estimated separately for each abnormality as well as for normal versus abnormal scans.

\absdiv{Results}
On the CQ2000 dataset, the deep learning system demonstrated an AUC of $0.92$(CI $0.91$-$0.94$) for detection of abnormal scans, and AUC (CI) of $0.96$($0.94$-$0.98$), $0.96$($0.94$-$0.98$), $0.95$($0.87$-$1$), $0.95$($0.92$-$0.98$), $0.93$($0.90$-$0.96$), $0.89$($0.83$-$0.94$), $0.91$($0.87$-$0.96$), $0.94$($0.93$-$0.96$), $0.98$($0.97$-$1$) for the detection of blunted costophrenic angle, cardiomegaly, cavity, consolidation, fibrosis, hilar enlargement, nodule, opacity and pleural effusion. The AUCs were similar on the larger CQ100k dataset except for detecting normals where the AUC was $0.86$($0.85$-$0.86$).

\absdiv{Interpretation}
Our study demonstrates that a deep learning algorithm trained on a large quantity of well-labelled data can accurately detect multiple abnormalities on chest X-rays. As these systems further increase in accuracy, the feasibility of applying deep learning to widen the reach of chest X-ray interpretation and improve reporting efficiency will add tremendous value in radiology workflows and public health screenings globally.
\end{abstract}

\section{Introduction}

Chest X-rays are the most commonly ordered diagnostic imaging tests, with millions of X-rays performed globally every year\cite{england2016diagnostic}. While the chest X-ray is frequently performed, interpreting a chest X-ray is one of the most subjective and complex of radiology tasks, with inter-reader agreement varying from a kappa value of $0.2$ to $0.77$, depending on the level of experience of the reader, the abnormality being detected and the clinical setting\cite{moncada2011reading,hopstaken2004inter,moifo2015inter,sakurada2012inter}. Due to their wide availability and affordability, chest X-rays are performed all over the world, including remote areas with few or no radiologists. In some parts of the world, digital chest X-ray machines are more widely available than personnel sufficiently trained to report the X-rays they generated\cite{crisp2014global}. If automated detection can be applied in low-resource settings as a disease screening tool, the benefits to population health outcomes globally could be significant. An example of the use of chest X-rays as a screening tool is in tuberculosis screening, where chest X-rays, in the hands of expert readers, are more sensitive than clinical symptoms for the early detection of tuberculosis\cite{world2016chest}.

Over the last few years, there has been increasing interest in the use of deep learning algorithms to assist with abnormality detection on medical images\cite{gulshan2016development,esteva2017dermatologist,chilamkurthy2018development}. This is a natural consequence of the rapidly growing ability of machines to interpret natural images and detect objects in them. On chest X-rays in particular, there have been a series of studies describing the use of deep learning algorithms to detect various abnormalities\cite{rajpurkar2017chexnet,hwang2019development}. Most of these have been limited by the lack of availability of large, high-quality datasets with the largest published work describing an algorithm that has been trained on 224,316 X-rays \cite{irvin2019chexpert}, a relatively small number considering that the majority of chest X-rays are normal, abnormal X-rays are less common and specific abnormalities being rarer still. The previously published work on deep learning for chest X-ray abnormality detection has not made a distinction between the diagnosis of `diseases' and the detection of `abnormal findings'. Our approach was to focus on the detection of abnormalities on the X-ray that can be detected visually by an expert without the benefit of the clinical history. This would allow the system to be applied across geographies and differing disease prevalence.

In this paper, we describe the training and radiologist validation of a deep learning system to detect and identify chest X-ray abnormalities. We trained the system on 2.3 million chest X-rays and tested it against the majority vote of a panel of 3 radiologist reads on an independent dataset(CQ2000) containing 2000 studies. We also validated it on another dataset(CQ100k) containing 100,000 scans where the gold standard is the radiologist report. A point to note is that the CQ100k is similar in order of size to the entire training datasets of most other published studies. Abnormalities on chest X-rays range from very small lesions to diffuse abnormalities that cover large parts of the lung. The optimal deep learning model architecture differs based on the abnormality being detected; hence, we developed and tested a system that uses an individual algorithm for each abnormality.

\section{Methods}
\subsection{Datasets}
\label{sec:datasets}
A dataset of 2.5 million anonymized chest X-rays was collected retrospectively from around 45 centres across the world and contained X-rays that were acquired in PA, AP, supine or lateral views. Additionally, the variation in size, resolution and quality of X-rays was significant. The centres from which this data was collected were a combination of in-hospital and out-patient radiology centers. Each of the X-rays had a corresponding radiology report associated with it.

Of the 2.5 million X-rays collected, 100,000 X-rays of randomly chosen 89,354 patients(CQ100k dataset) were used for validation and X-rays from rest of the patients(development dataset) were used for algorithm development. We excluded lateral chest X-rays and X-rays of patients younger than 14 years from the CQ100k dataset. This dataset was not used during the algorithm development process.

A further validation dataset(CQ2000) was created from a combination of out-patient and in-hospital X-rays from three Columbia Asia Hospitals at Kolkata, Pune and Mysore in India. There was no overlap between these centers and those used to obtain the development dataset or CQ100k dataset. The data was transferred considering all security and privacy aspects. Before the data was shared, it was stripped off all patient identifiable information and the processes were carried out in a controlled environment compliant with all Indian IT laws and the Health Insurance Portability and Accountability Act(HIPAA). As with the development and CQ100k datasets, each X-ray in the CQ2000 dataset had a corresponding clinical radiology report available. These reports were used for both dataset selection and establishing the gold standard as described in section \ref{sec:sor}. 

\begin{table}
	\centering
	\begin{tabularx}{0.8\textwidth}{l X X}
	\toprule
	\textbf{X-Ray Pool} & \textbf{Center} & \textbf{Date} \\
	\midrule
	\multirow{3}{*}{Pool 1}&CAH-Kolkata&16-08-2017 to 31-08-2017 \\ 
	&CAH-Pune & 07-09-2017 to 23-09-2017 \\ 
	
	&CAH-Mysore & 10-07-2017 to 25-07-2017 \\ 
	\midrule
	Pool 2&CAH-Pune&01-02-2013 to 31-07-2013 \\ 
	\bottomrule
	\end{tabularx}
	\bigskip
	\caption{Source of X-rays used for the study}
	\label{table:xraysource}
\end{table}

For creating the CQ2000 dataset, data was first collected in two pools as described in table \ref{table:xraysource}. Chest X-rays from each pool were filtered at source as follows: all PA and AP view chest X-rays where a corresponding radiologist report was available were selected. X-rays from patients younger than 14 years of age, X-rays taken in the supine position - from bedside or portable X-ray machines - were excluded. As a result, the final CQ2000 dataset did not contain any X-rays with visible intravenous lines, tubes, cathereters, ECG leads or any implanted medical devices such as pacemakers. A set of 1000 X-rays(Set 1) was then selected randomly from Pool 1. A second set of 1000 X-rays(Set 2) were randomly sampled from Pool 2 to include around 80 examples of each abnormality listed in table \ref{table:tagdefs}. A natural language processing(NLP) algorithm was used to parse the X-ray radiology reports and implement the exclusions listed above. A second NLP algorithm was used to detect various abnormalities from radiology reports in Pool 2 to automate the sampling for Set 2. This NLP algorithm is same as the one that was used to establish the gold standard on the development dataset and the CQ100k dataset as described in section \ref{sec:sor}

\begin{table}
	\centering
	\begin{tabularx}{\textwidth}{p{5.3em} XX XX}
	\toprule
	\textbf{Finding} &
	\textbf{Definition for tag extraction from \newline radiology reports} &
	\textbf{Definition for radiologist review} \\

	\midrule
	Normal &
	`No abnormality detected' / `Normal' &
	Normal X-ray\\
	
	\midrule
	Blunted CP angle &
	Blunted Costophrenic(CP) angle &
	CP angle blunted/obscured: pleural effusion/pleural thickening\\

	\midrule
	Cardiomegaly &
	Cardiomegaly &
	Cardiothoracic ratio > 0.5\\

	\midrule
	Cavity &
	Pulmonary cavity &
	Pulmonary cavity\\

	\midrule
	Consolidation &
	Consolidation/ pneumonia/ \newline air-bronchogram &
	Pulmonary consolidation\\

	\midrule
	Fibrosis &
	Fibrosis &
	Lung fibrosis/ interstitial fibrosis/ fibrocavitary lesion\\

	\midrule
	Hilar \newline prominence &
	Hilar enlargement/ prominent hilum/ hilar lymphadenopathy &
	Enlarged hilum/ prominent hilum/ hilar lymphadenopathy\\

	\midrule
	Nodule &
	Nodule &
	Any abnormal, small well-defined opacities in the lung fields smaller than 3 cm in diameter \\

	\midrule
	Opacity &
	Lung opacity/ opacities/ shadow/ density/ infiltrate, consolidation/ mass/ nodule/ pulmonary calcification/ fibrosis &
	Any lung opacity/ multiple opacities including: infiltrate/ consolidation/ mass/ nodule/ pulmonary calcification/ fibrosis \newline Note: pleural abnormalities not included under this tag\\

	\midrule
	Pleural \newline Effusion &
	Pleural Effusion &
	Pleural Effusion \\

	\bottomrule
	\end{tabularx}
	\smallskip
	\caption{Abnormality definitions}
	\label{table:tagdefs}
\end{table}

\subsection{Establishing gold standards}
\label{sec:sor}
For the development and CQ100k datasets, we considered clinical reports written by radiologists as the gold standard. Since the reports were acquired in a free text format, we used a custom NLP algorithm to parse these and detect the following target chest X-ray findings: blunted costophrenic angle, cardiomegaly, cavity, consolidation, fibrosis, hilar enlargement, nodule, opacity and pleural effusion. The NLP algorithm was constructed using a thoracic imaging glossary \cite{hansell2008fleischner}, curated by a panel of radiologists and tailored to be consistent with the abnormality definitions given in table \ref{table:tagdefs}. It is then combined with standard NLP tools to manage typographic errors, detect negations and identify synonyms. To verify that the automatically inferred findings are consistent with tag definitions(table \ref{table:tagdefs}), we validated the NLP algorithm on a separate independent dataset of 1930 chest X-rays against an expert reader who was provided with the tag definitions (table \ref{table:tagdefs}), original reports and the corresponding X-rays. These expert readers were blinded to the algorithm output. Results from this validation are presented in table \ref{table:tagexacc}.

Six certified radiologists with 3-15 years of radiology experience served as readers for the X-rays in CQ2000 dataset. For the sake of consistency, the readers used the definitions in table \ref{table:tagdefs} as a frame of reference for labelling the images. The X-rays were randomly divided among the readers and each X-ray was read by 2 readers independently. When there was no unanimous agreement between the two readers for a particular X-ray, original clinical report of the X-ray was used as the tie-breaker to establish the gold standard.

\begin{figure}
	\centering
	\begin{subfigure}[t]{0.90\textwidth}
            \centering
           \includegraphics[width=0.9\linewidth]{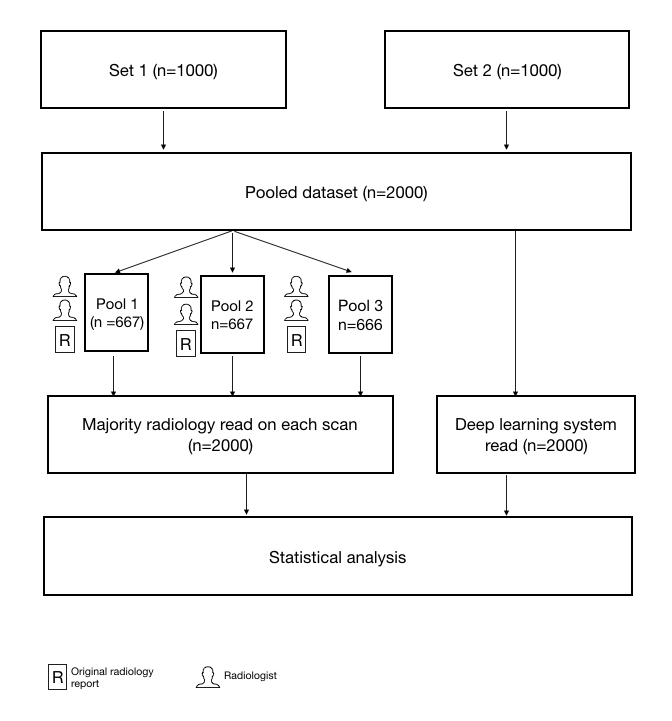}
    \end{subfigure}
    \caption{Study Design.}
    \label{fig:studydesign}
\end{figure}

\subsection{Algorithm Development}
We used the X-rays from the development set and the labels automatically inferred from the corresponding radiology reports for developing the algorithms. The algorithms can identify normal X-rays, and the following chest X-ray findings: `blunted CP angle', `cardiomegaly', `cavity', `consolidation', `fibrosis', `hilar enlargement', `nodule', `opacity', `pleural effusion'. 

We used deep learning to train a type of neural networks called convolutional neural networks(CNNs). The specific architectures that form the basic blocks in the systems that detect individual abnormalities are versions of resnets\cite{he2016deep} with sqeeze-excitation modules\cite{hu2018squeeze}. The vanilla versions of these architectures are modified to process information at a significantly higher resolution. The chest X-rays in the development set varied considerably in size, resolution and quality. Before they are presented to the networks, the X-rays were down-sampled and resized to a standard size and a set of image normalization techniques was applied to reduce source-dependent variation. Additionally, a number of abnormality-specific data augmentation techniques were used. The aim of data augmentation is to generate a dataset that can be used to train networks that are unaffected by variability in X-ray machine manufacturer, model, voltage, exposure and other parameters that vary from center to center. All classification networks that build up the individual abnormality detection systems are pre-trained on the task of separating chest X-rays from X-rays of other body parts rather than the popular ImageNet pre-training. This step is aimed at making use of the super-set consisting of all the X-rays. We observed improved model convergence and incremental gains in generalization performance when compared to ImageNet pre-training.

Model ensembling\cite{krizhevsky2012imagenet} is a simple way of improving generalization performance by combining the predictions produced by a set of models. We trained multiple networks to detect each abnormality. These networks differ with respect to the architecture used, model initialization conditions and the distribution of the training dataset. A subset of these models was selected using various heuristics\cite{caruana2004ensemble} and a majority ensembling scheme is used to combine the predictions of these selected models to make a decision about the presence or absence of a particular abnormality. 

\subsection{Validation of the algorithms}
When presented with a chest X-ray, the set of trained neural networks produce one real-valued confidence number between 0 and 1 per abnormality indicating its presence or absence. Gold standards were established on the two validation sets - CQ2000 and CQ100k as described in section \ref{sec:sor}. Algorithms were assessed independently for each finding. 
For both the validation datasets, we computed the Receiver operating characteristic(ROC) curves\cite{hanley1982meaning} for each finding and chose two operating points - one at a high sensitivity(approximately 0.9) and another at a high specificity(approximately 0.9). Areas under the ROC curves(AUCs) and sensitivities and specificities at the selected operating points on the ROC curves were used to assess the performance of the algorithms on individual target findings.

\subsection{Statistical Analysis}
As we are evaluating the algorithms independently for each finding, this study is a combination of individual diagnostic tests for each of the target findings. We used the normal approximation for calculating the sample sizes for proportions and the method outlined by Hanley and McNeil\cite{hanley1982meaning} for calculating the AUCs. For calculating the sample sizes of a particular diagnostic test, sensitivity was chosen over specificity as a false negative result is considered a worse outcome than a false positive in preliminary diagnostic investigations \cite{schwartz2000us}. To establish sensitivity at an expected value of $80\%$ at $10\%$ precision and $95\%$ CI, the number of positive scans to be read is appproximately 80. 
In a randomly selected sample, the prevalence of some of our target abnormalities are as low as $1\%$. To establish AUC at $5\%$ precision and a $95\%$ CI for such findings in a random population, the number of samples to be read is \~15000.

Of the two validation datasets, the CQ100k dataset was sampled randomly and satisfies all the sample size requirements to validate the algorithms. However, on the CQ2000 dataset, due to constraints on the number of radiologist reads required, we used the enrichment strategy detailed in section \ref{sec:datasets} to meet these requirements.

The $95\%$ confidence intervals for sensitivity and specificity of a particular target abnormality detection algorithm at the two operating points selected(one at high sensitivity, another at high specificity) is calculated to be 'exact' using Clopper-Pearson intervals.\cite{clopper1934use} The $95\%$ confidence intervals for AUCs were calculated following the distribution-based detailed by Hanley and McNeil.\cite{hanley1982meaning} On the CQ2000, We calculated the concordance between the two radiologist reads on a given X-ray using the Cohen's \(\kappa\) statistic and the percentage agreement. The percentage agreement between two reads is also equivalent to the proportion of X-rays that didn't need a clinical report tie-breaker to establish the gold standard.

\section{Results}
Basic demographics and the prevalences of each target finding are summarized in table \ref{table:demographics}. In the CQ2000 dataset, 1342 out of 2000 X-rays were abnormal, with the most frequent abnormalities being `opacity' and `cardiomegaly'. There were insufficient X-rays in CQ2000 with ‘cavity’ to confidently calculate the accuracy of the deep learning system in identifying this abnormality. In CQ100k dataset, there were 34433 abnormal scans out of 100000 and there were sufficient scans for all abnormalities to confidently estimate the accuracies.

Achieving a high accuracy on report parsing enabled the use of a large number of X-rays to train the deep learning algorithms. Abnormality extraction accuracy from radiology reports versus manual extraction by a single reader is summarized in table \ref{table:tagexacc}. The NLP algorithm was able to detect normal X-ray reports with a sensitivity of 0.94 and a specificity of 1 versus the expert reader. For detection of individual abnormalities from reports, sensitivity varied from 0.93 for pleural effusion to 1 for cavity; specificity varied from 0.92 for opacity to 0.99 for fibrosis. 

Inter-reader concordance is described in table \ref{table:concordance}. Concordance was highest on detection of abnormal X-rays (inter-reader agreement $85\%$, Cohen’s kappa 0.6, Fleiss’ kappa 0.56) and on the specific abnormalities pleural effusion (inter-reader agreement $85\%$, Cohen’s kappa 0.6, Fleiss’ kappa 0.56), cardiomegaly (inter-reader agreement $85\%$, Cohen’s kappa 0.6, Fleiss’ kappa 0.56).

\begin{figure}
	\begin{subfigure}{0.32\textwidth}
	\includegraphics[width=\linewidth]{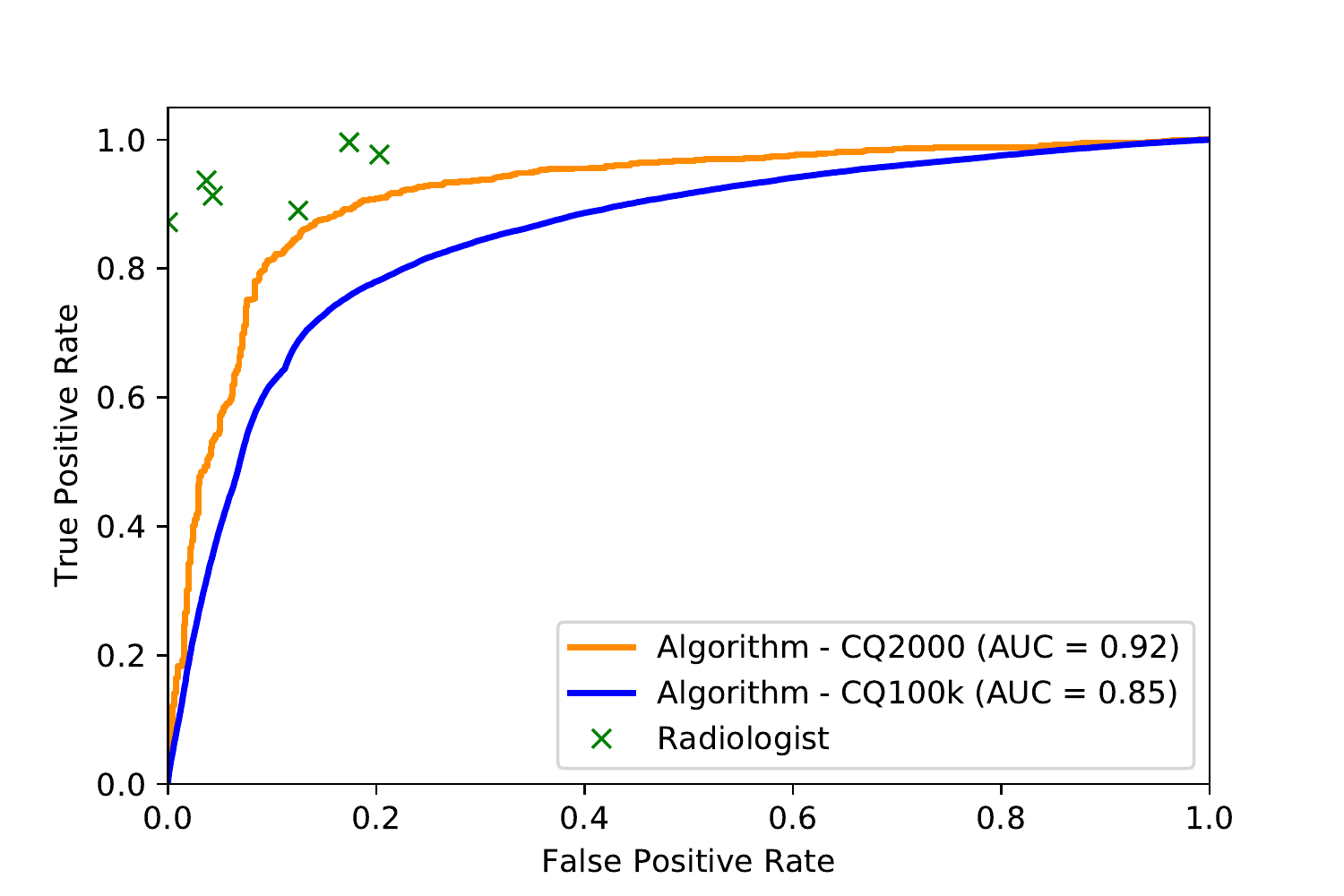}
	\caption*{\small Abnormal}
	\end{subfigure}
	\hfill
	\begin{subfigure}{0.32\textwidth}
	\includegraphics[width=\linewidth]{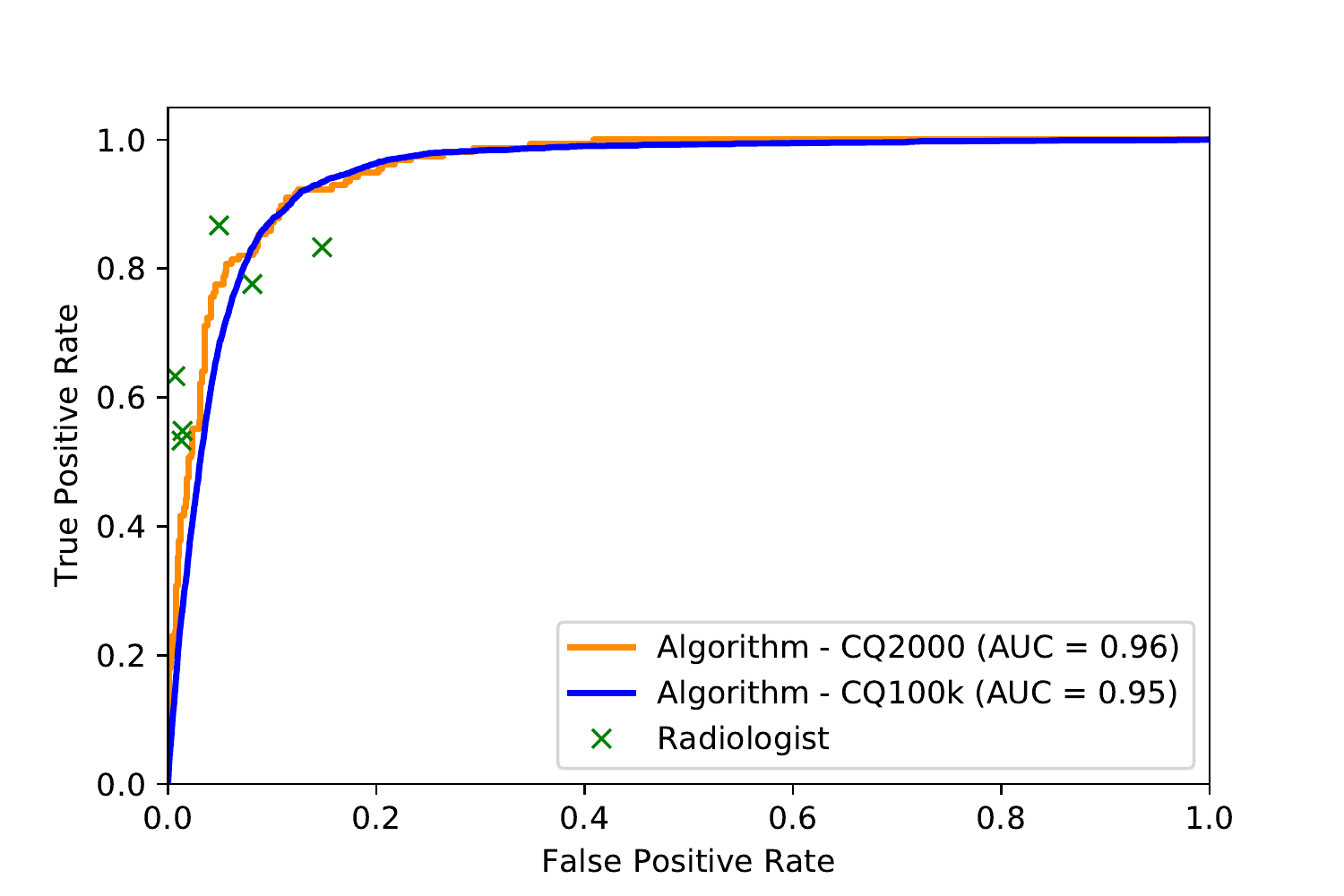}
	\caption*{\small Blunted CP angle}
	\end{subfigure}
	\hfill
	\begin{subfigure}{0.32\textwidth}
	\includegraphics[width=\linewidth]{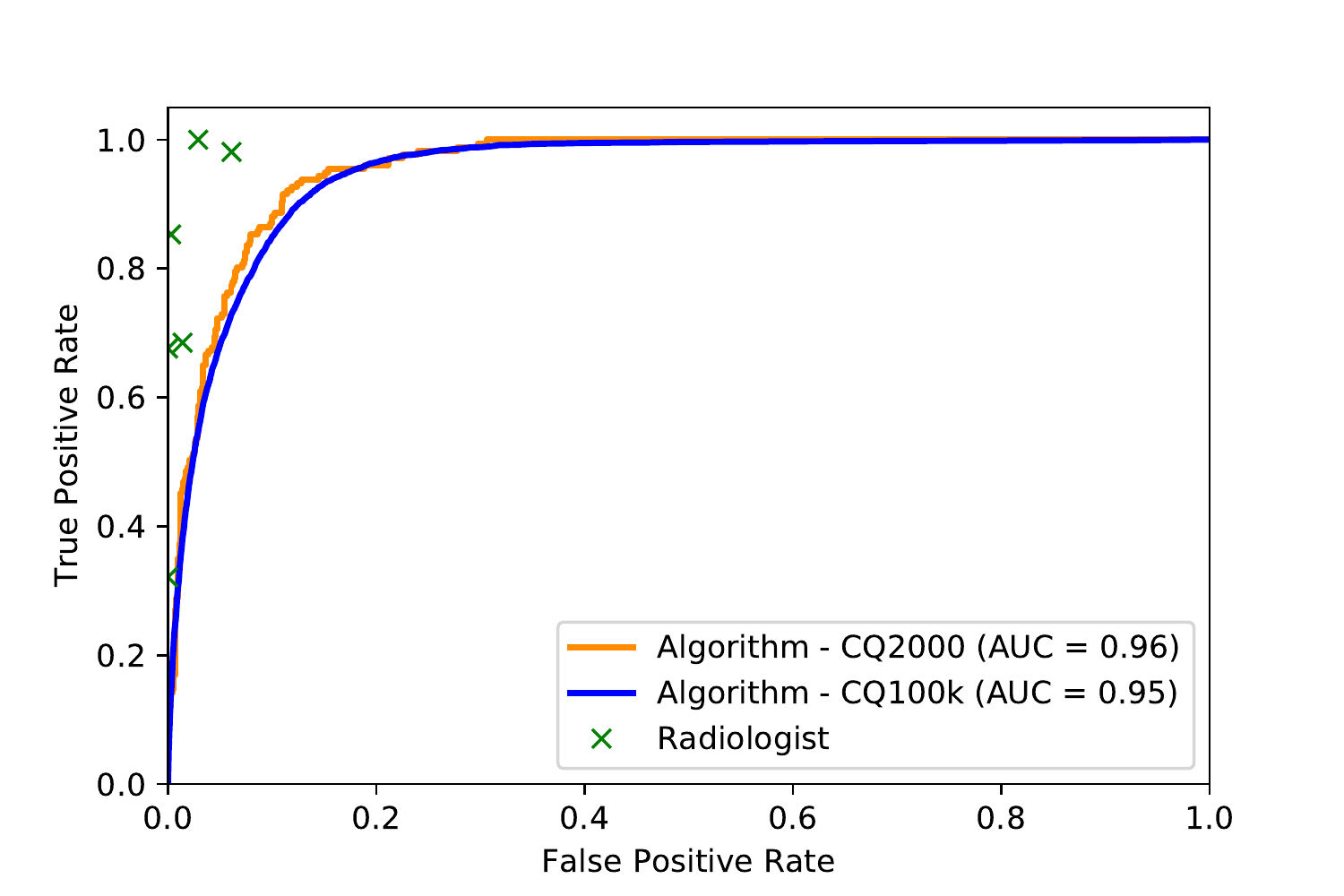}
	\caption*{\small Cardiomegaly}
	\end{subfigure}
	\bigskip

	\begin{subfigure}{0.32\textwidth}
	\includegraphics[width=\linewidth]{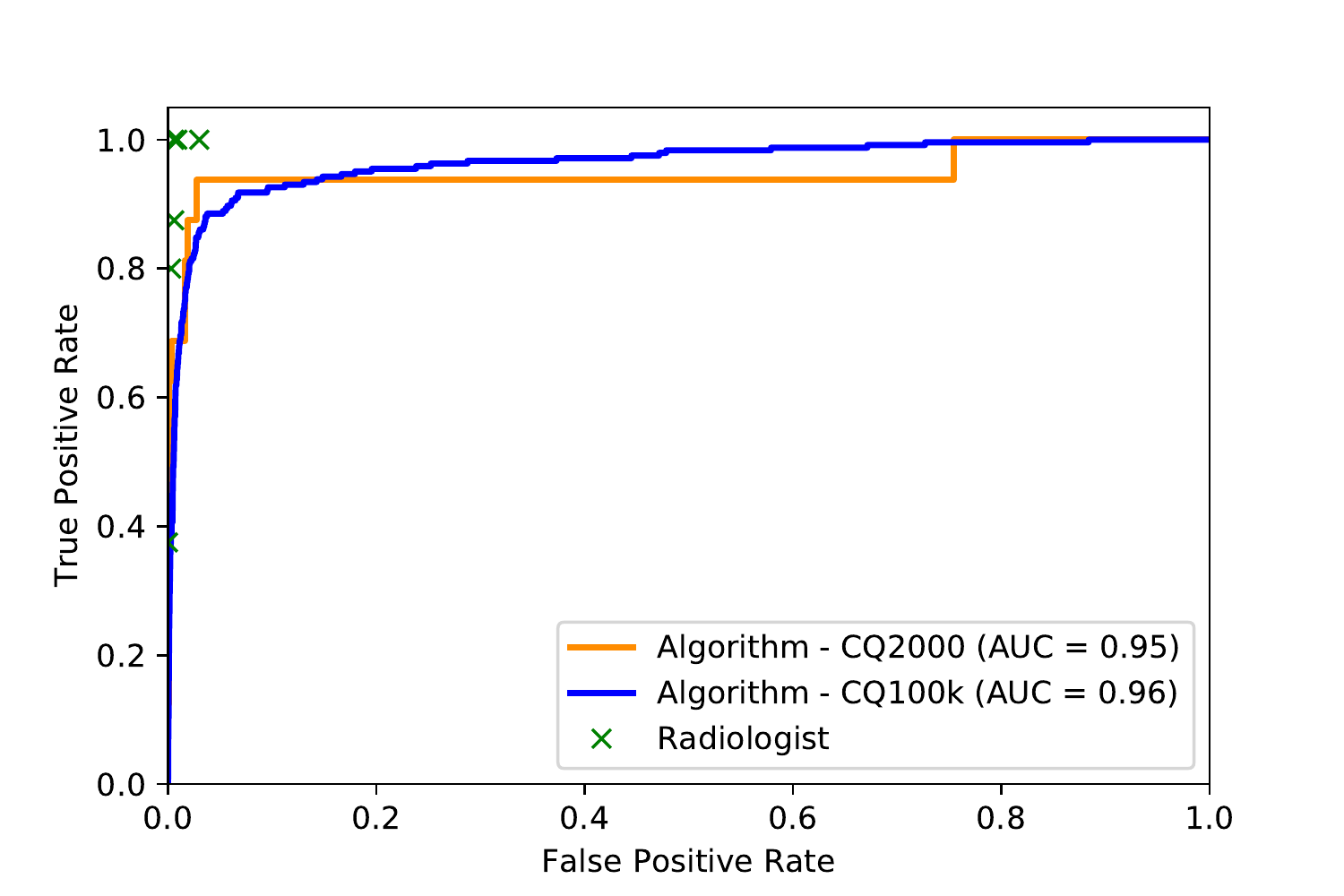}
	\caption*{\small Cavity}
	\end{subfigure}
	\hfill
	\begin{subfigure}{0.32\textwidth}
	\includegraphics[width=\linewidth]{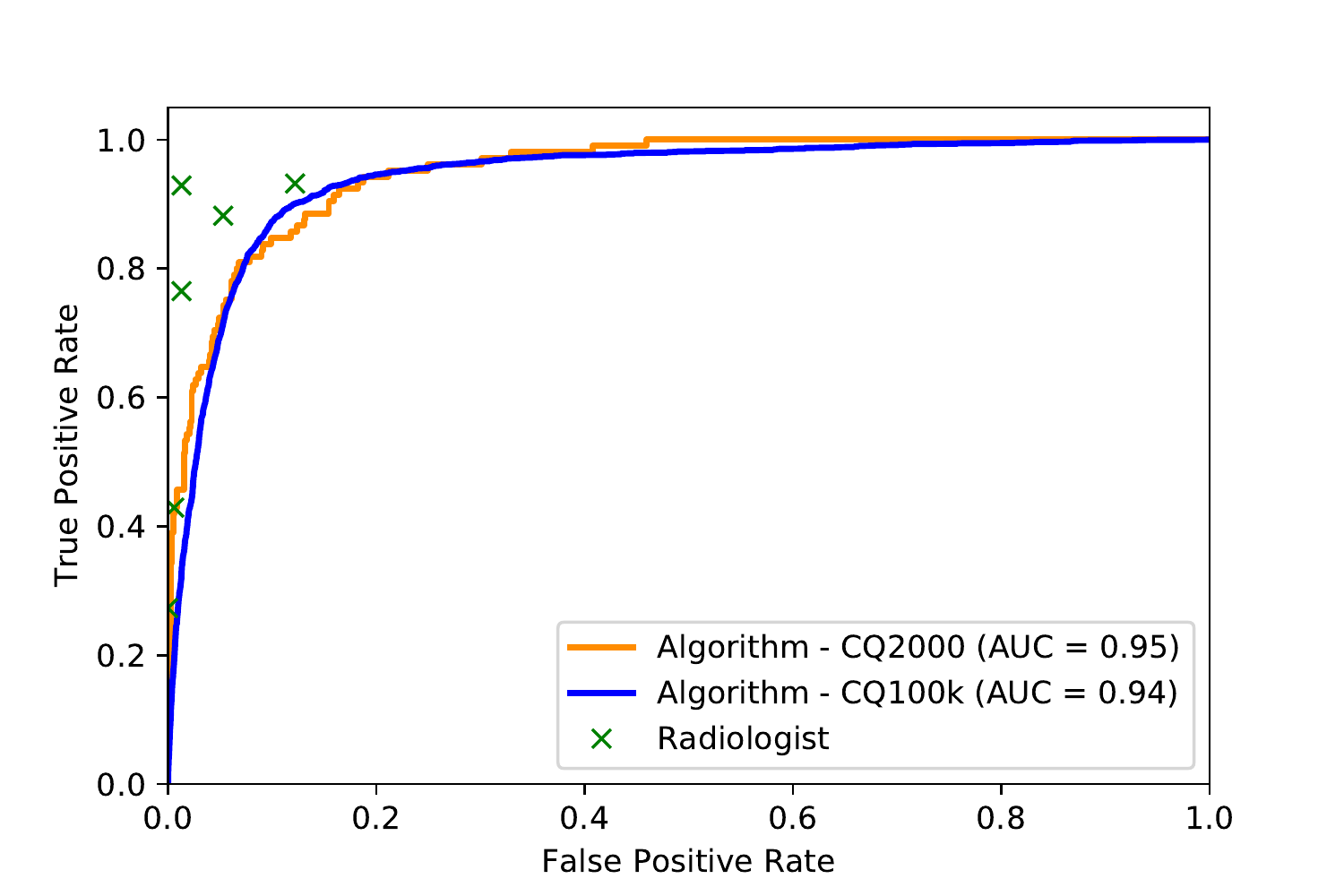}
	\caption*{\small Consolidation}
	\end{subfigure}
	\hfill
	\begin{subfigure}{0.32\textwidth}
	\includegraphics[width=\linewidth]{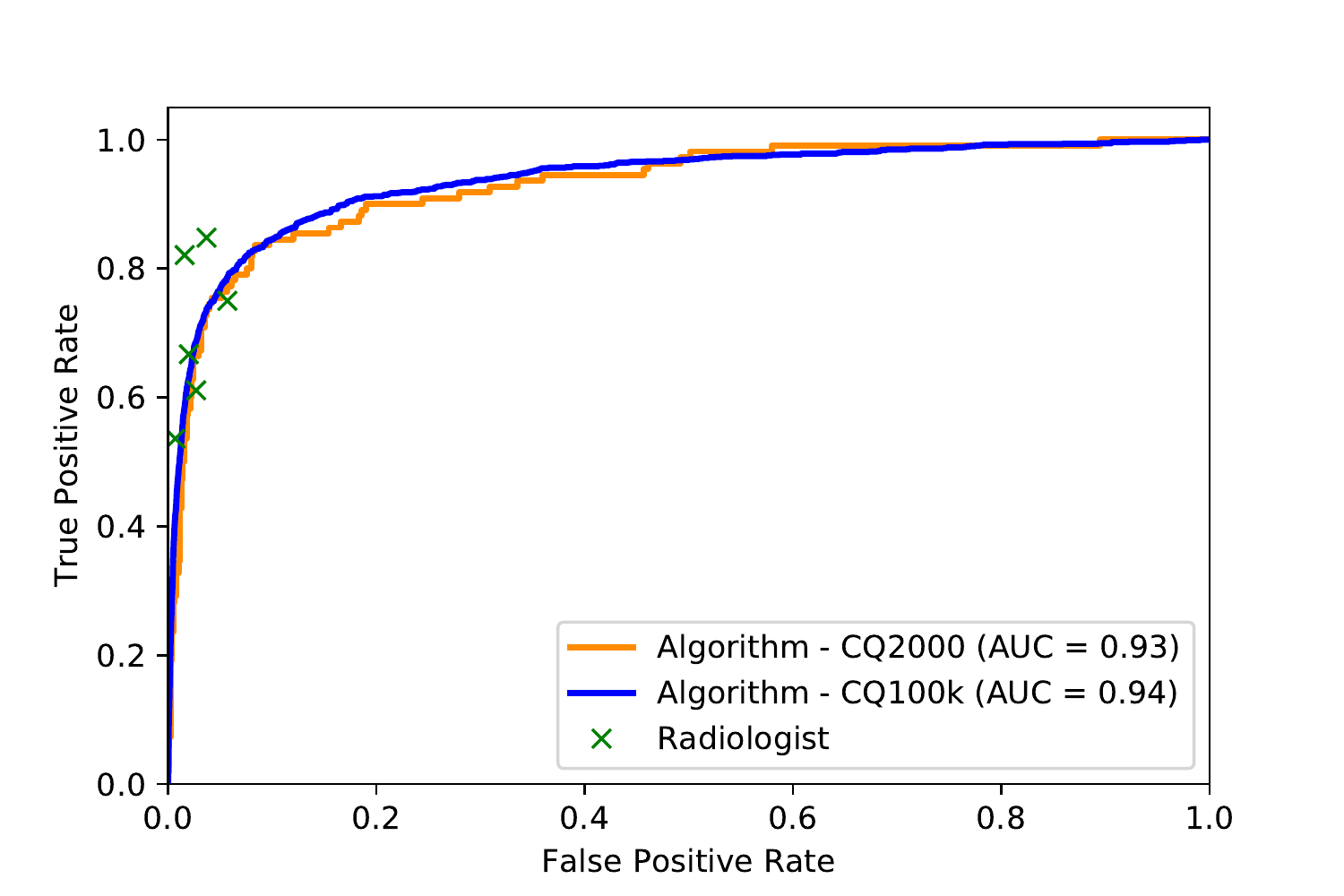}
	\caption*{\small Fibrosis}
	\end{subfigure}
	\bigskip
	
	\begin{subfigure}{0.32\textwidth}
	\includegraphics[width=\linewidth]{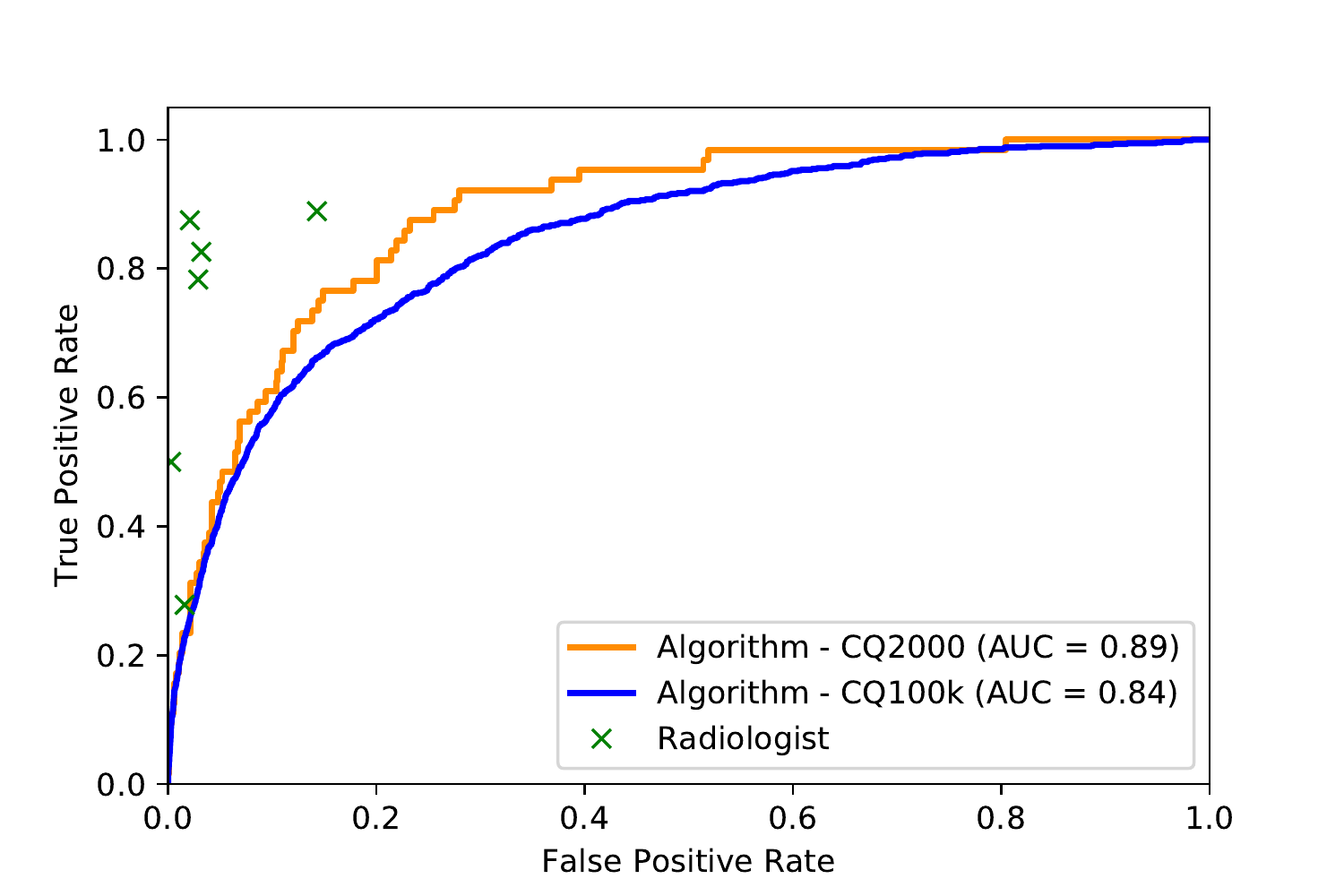}
	\caption*{\small Hilar Prominence}
	\end{subfigure}
	\hfill
	\begin{subfigure}{0.32\textwidth}
	\includegraphics[width=\linewidth]{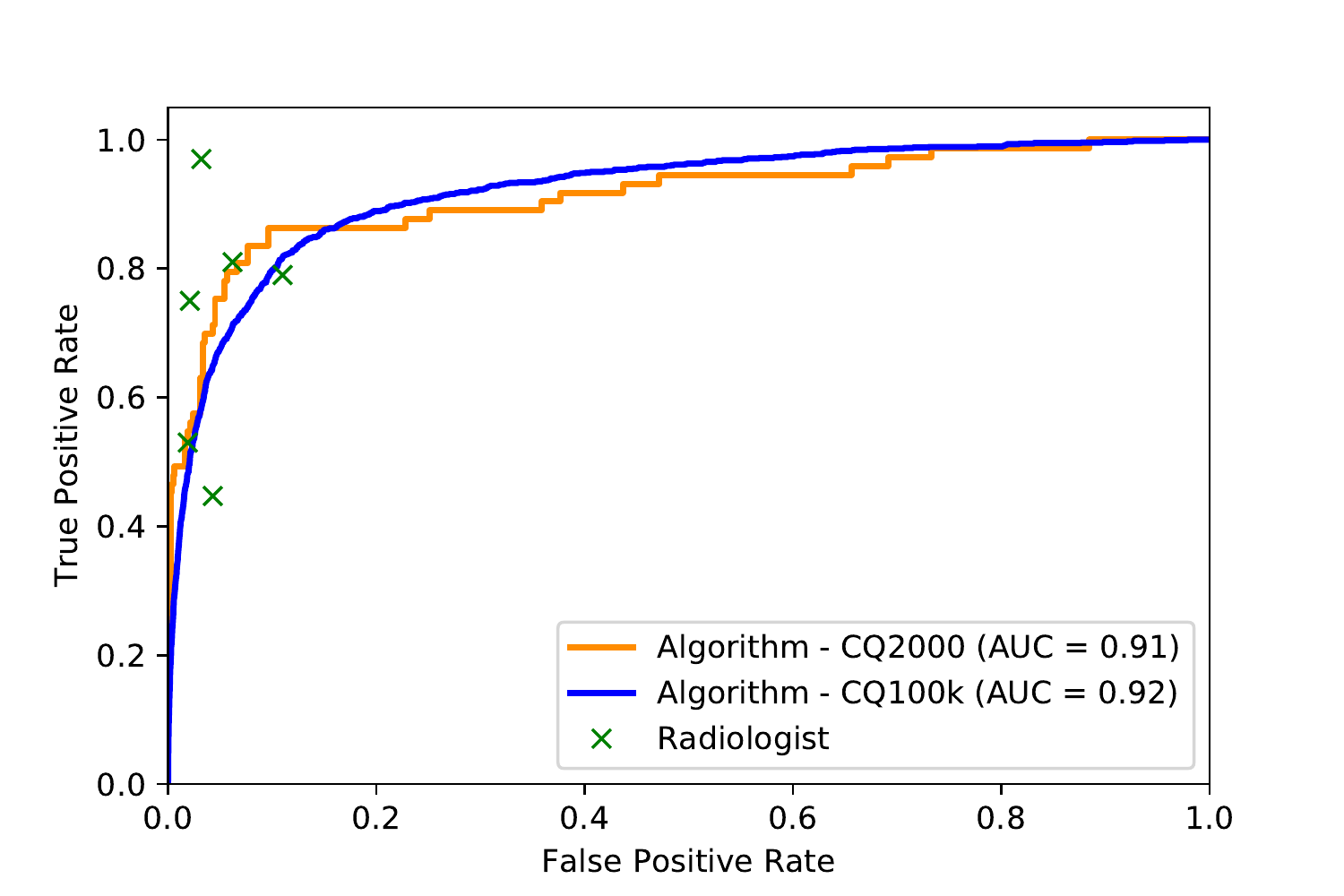}
	\caption*{\small Nodule}
	\end{subfigure}
	\hfill
	\begin{subfigure}{0.32\textwidth}
	\includegraphics[width=\linewidth]{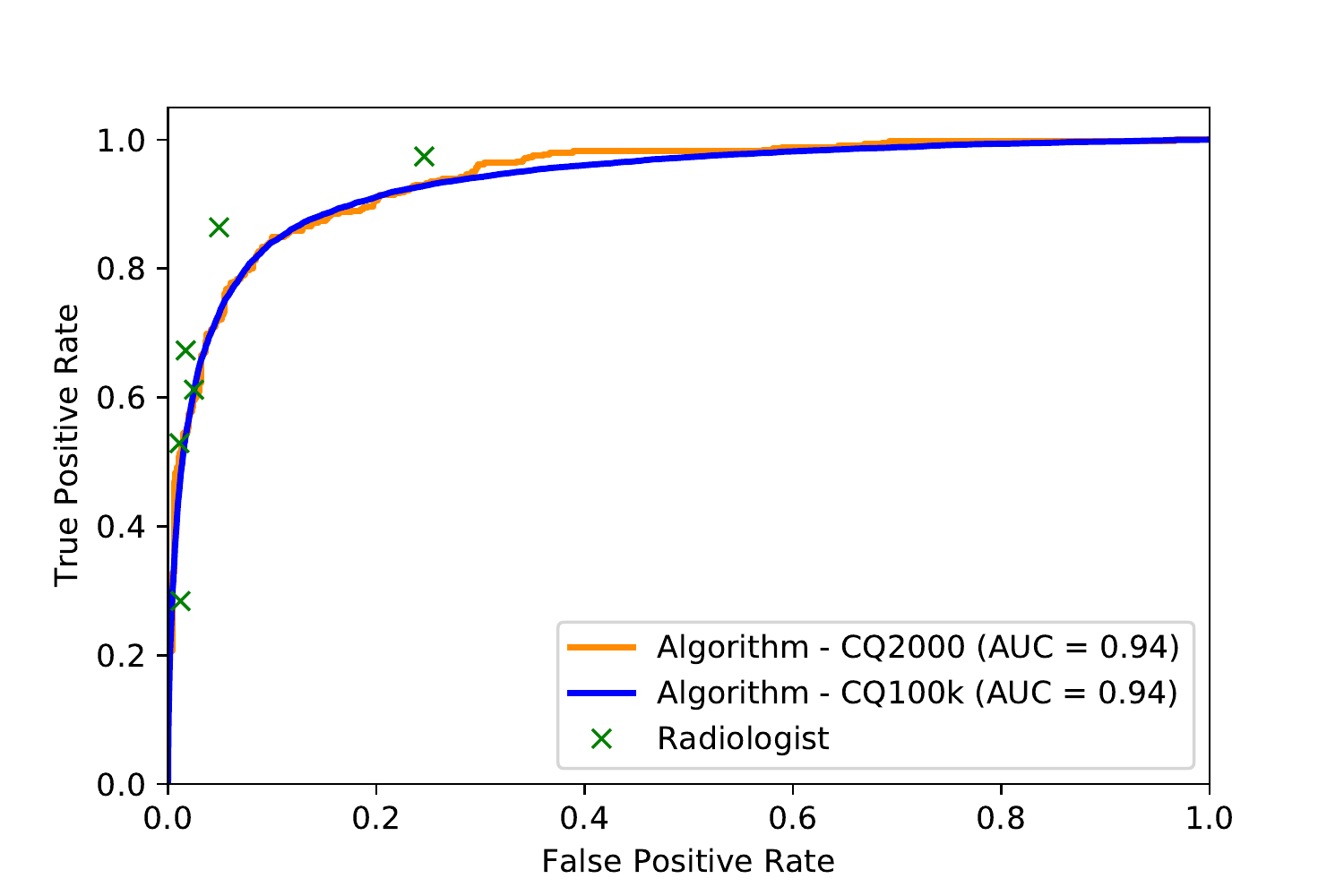}
	\caption*{\small Opacity}
	\end{subfigure}
	\bigskip

	\begin{subfigure}{0.32\textwidth}
	\includegraphics[width=\linewidth]{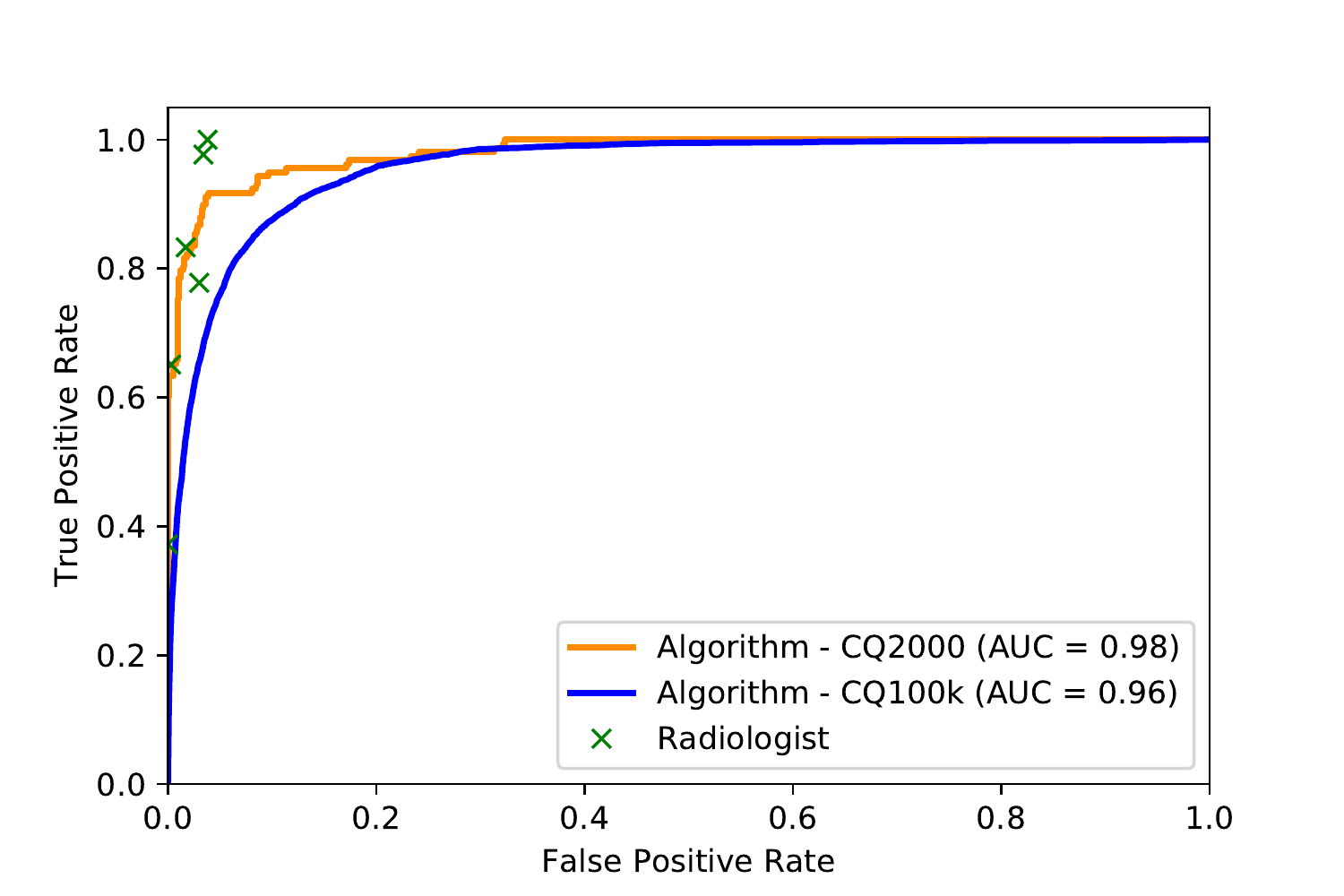}
	\caption*{\small Pleural Effusion}
	\end{subfigure}
	\bigskip
	\caption{AUC curves for all abnormalities versus a 3-radiologist majority for CQ2000(red) and for CQ100k(blue), with reader performance marked.(green)}
    \label{fig:cxr_auc}

\end{figure}

The deep learning system accuracy at identifying each of the 10 abnormalities is listed in table \ref{table:results}. Figure \ref{fig:cxr_auc} shows ROC curves for each abnormality on the two datasets. Individual radiologist sensitivity and specificity for the 6 radiologists on the CQ2000 dataset is marked on each plot. In most cases, individual radiologist sensitivity and specificity was marginally above the ROC curve, with exceptions for pleural effusion, cardiomegaly and opacity where algorithm performance was equal to the performance of some individuals.

\begin{table}
	\centering
	\bigskip
	\begin{tabularx}{\textwidth}{p{17em} X X X}
	\toprule

	\textbf{Characteristic} & \textbf{CQ2000 Set1} &
	\textbf{CQ2000 Set2} & \textbf{CQ100k} \\

	\midrule

	No. of scans & \(1000\) & \(1000\) & \(100000\) \\

	No. of readers per scan & \(3\) & \(3\) & \(1\) \\

	\midrule
	\multicolumn{4}{l}{\textsc{Patient Demographics}} \\

	\multicolumn{4}{l}{Age} \\
	\hspace*{1em} No. of scans for which age was known & \(803\) & \(955\) & \(99920\)\\
	\hspace*{1em} Mean & \(48.04\) & \(50.61\) & \(44.92\) \\
	\hspace*{1em} Standard deviation & \(18.69\) & \(17.93\) & \(17.24\) \\
	\hspace*{1em} Range & \(16 - 95\) & \(16 - 100\) & \(16 - 117\) \\

	No. of females / No. of scans for which sex was known (percentage)  & 
	\(324 / 803\) \newline \small (\(40.3\)\%) &
	\(265 / 1000\) \newline \small (\(26.5\)\%) &
	\(37799/ 99920\) \newline \small (\(37.8\)\%) \\

	\midrule
	\multicolumn{4}{l}{\textsc{Prevalence}} \\
	\multicolumn{4}{l}{No. of scans (percentage) with } \\
	\hspace*{1em} No abnormality detected(Normal)
	& \(440\) &
    \(177\) &
    \(65567\) \\

	\hspace*{1em} Blunted CP angle
	& \(35\) &
    \(123\) &
    \(2853\)\\

	\hspace*{1em} Cardiomegaly
	& \(61\) &
    \(116\) &
    \(4636\)\\

	\hspace*{1em} Cavity
	& \(1\) &
    \(15\) &
    \(205\)\\

	\hspace*{1em} Consolidation
	& \(13\) &
    \(94\) &
    \(2007\)\\

	\hspace*{1em} Fibrosis
	& \(13\) &
    \(106\) &
    \(1174\)\\

	\hspace*{1em} Hilar enlargement
	& \(15\) &
    \(53\) &
    \(795\)\\

    \hspace*{1em} Nodule
	& \(6\) &
    \(45\) &
    \(1202\)\\

	\hspace*{1em} Opacity
	& \(104\) &
    \(386\) &
    \(12746\)\\

	\hspace*{1em} Pleural Effusion
	& \(36\) &
    \(123\) &
    \(4130\)\\

	\bottomrule
	\end{tabularx}
	\bigskip
	\caption{Demographics of the study population}
	\label{table:demographics}{}
\end{table}

\begin{table}
	\begin{subtable}{\textwidth}
		\centering

		\begin{tabularx}{\textwidth}{ p{15em} p{5em} YY }
		\toprule

		\textbf{Finding} & \textbf{\#Positives} &
		\textbf{Sensitivity \newline \small($95\%$ CI)} &
		\textbf{Specificity \newline \small($95\%$ CI)} \\

		\midrule
		Normal(No abnormality detected) & \(105\) &
		\(0.9429\) \newline \tiny{(\(0.8798\)-\(0.9787\))} &
		\(1.0000\) \newline \tiny{(\(0.9959\)-\(1.0000\))} \\

		\midrule
		Blunted CP angle & \(146\) &
		\(0.9795\) \newline \tiny{(\(0.9411\)-\(0.9957\))} &
		\(0.9824\) \newline \tiny{(\(0.9712\)-\(0.9901\))} \\

		\midrule
		Cardiomegaly & \(125\) &
		\(0.9920\) \newline \tiny{(\(0.9562\)-\(0.9998\))} &
		\(0.9985\) \newline \tiny{(\(0.9916\)-\(1.0000\))} \\

		\midrule
		Cavity & \(30\) &
		\(1.0000\) \newline \tiny{(\(0.8843\)-\(1.0000\))} &
		\(0.9856\) \newline \tiny{(\(0.9759\)-\(0.9921\))} \\

		\midrule
		Consolidation & \(161\) &
		\(0.9876\) \newline \tiny{(\(0.9558\)-\(0.9985\))} &
		\(0.9761\) \newline \tiny{(\(0.9634\)-\(0.9854\))} \\

		\midrule
		Fibrosis & \(124\) &
		\(0.9839\) \newline \tiny{(\(0.9430\)-\(0.9980\))} &
		\(0.9931\) \newline \tiny{(\(0.9851\)-\(0.9975\))} \\

		\midrule
		Hilar Enlargement & \(289\) &
		\(0.9689\) \newline \tiny{(\(0.9417\)-\(0.9857\))} &
		\(0.9732\) \newline \tiny{(\(0.9585\)-\(0.9838\))} \\

		\midrule
		Nodule & \(92\) &
		\(0.9783\) \newline \tiny{(\(0.9237\)-\(0.9974\))} &
		\(0.9660\) \newline \tiny{(\(0.9519\)-\(0.9770\))} \\

		\midrule
		Opacity & \(612\) &
		\(0.9608\) \newline \tiny{(\(0.9422\)-\(0.9747\))} &
		\(0.9251\) \newline \tiny{(\(0.8942\)-\(0.9492\))} \\

		\midrule
		Pleural Effusion & \(246\) &
		\(0.9309\) \newline \tiny{(\(0.8917\)-\(0.9592\))} &
		\(0.9602\) \newline \tiny{(\(0.9436\)-\(0.9730\))} \\

		\midrule
		Total(all findings) & \(1930\) &
		\(0.9672\) \newline \tiny{(\(0.9584\)-\(0.9747\))} &
		\(0.9771\) \newline \tiny{(\(0.9736\)-\(0.9803\))} \\

		\bottomrule
		\end{tabularx}
		\smallskip
		\caption{Tag extraction accuracy: performance of the NLP algorithm in inferring findings from the reports.}
		\label{table:tagexacc}
	\end{subtable}
	\bigskip

	\begin{subtable}{\textwidth}
		\centering
		\begin{tabularx}{\textwidth}{p{15em} YY Y}
			\toprule

			\textbf{Finding} &
			\multicolumn{2}{c}{\textbf{Radiologist 1 \& 2}} &
			\textbf{All reads} \\

			&
			\small{Agreement \%} & \small{Cohen's \(\kappa\)} &
			\small{Fleiss' \(\kappa\)}
			\\

			\midrule
			Normal(No abnormality detected)&
			\(85.00\) & \(0.6049\) &
			\(0.5618\) \\

			\midrule
			Blunted CP angle&
			\(83.58\) & \(0.2968\) &
			\(0.3054\) \\

			\midrule
			Cardiomegaly&
			\(91.60\) & \(0.5333\) &
			\(0.5284\) \\

			\midrule
			Cavity&
			\(97.50\) & \(0.3824\) &
			\(0.4047\) \\

			\midrule
			Consolidation&
			\(88.28\) & \(0.3529\) &
			\(0.3397\) \\

			\midrule
			Fibrosis&
			\(89.40\) & \(0.3781\) &
			\(0.3495\) \\

			\midrule
			Hilar Enlargement&
			\(89.38\) & \(0.2630\) &
			\(0.2101\) \\

			\midrule
			Nodule&
			\(92.80\) & \(0.5471\) &
			\(0.5467\) \\

			\midrule
			Opacity&
			\(70.70\) & \(0.2306\) &
			\(0.1733\) \\

			\midrule
			Pleural Effusion&
			\(90.69\) & \(0.5341\) &
			\(0.5305\) \\

			\bottomrule
		\end{tabularx}
		\smallskip
		\caption{Concordance between the readers.}
		\label{table:concordance}
	\end{subtable}
	\caption{Reliability of the gold standard: Abnormality extraction accuracy from radiology reports versus manual extraction by a single reader is summarized in table \ref{table:tagexacc}. Inter-reader concordance is described in table \ref{table:concordance}}

\end{table}

\begin{table}
	\centering
	\begin{tabularx}{\linewidth}{p{8.5em} p{4.9em} p{4.9em} p{4.9em} p{4.9em}p{4.9em}}
	\toprule

	\textbf{Finding} & \textbf{AUC \newline \small ($95\%$ CI)} &
	\multicolumn{2}{p{10em}}{\textbf{High sensitivity operating point}} &
	\multicolumn{2}{p{10em}}{\textbf{High specificity operating point}} \\

	& &
	Sensitivity&
	Specificity&
	Sensitivity&
	Specificity \\

	\midrule
	Normal(No Abnormality Detected) &
	\(0.9227\) \newline \tiny{(\(0.9107\)-\(0.9348\))}  &
	\(0.9004\) \newline \tiny{(\(0.8816\)-\(0.9172\))}  &
	\(0.8146\) \newline \tiny{(\(0.7844\)-\(0.8450\))}  &
	\(0.7991\) \newline \tiny{(\(0.7738\)-\(0.8211\))}  &
	\(0.9012\) \newline \tiny{(\(0.8758\)-\(0.9229\))}  \\

	\midrule
	Blunted CP angle &
	\(0.9560\) \newline \tiny{(\(0.9333\)-\(0.9787\))}  &
	\(0.8974\) \newline \tiny{(\(0.8388\)-\(0.9402\))}  &
	\(0.8867\) \newline \tiny{(\(0.8636\)-\(0.9094\))}  &
	\(0.8718\) \newline \tiny{(\(0.8016\)-\(0.9147\))}  &
	\(0.9010\) \newline \tiny{(\(0.8777\)-\(0.9212\))}  \\

	\midrule
	Cardiomegaly &
	\(0.9577\) \newline \tiny{(\(0.9367\)-\(0.9786\))}  &
	\(0.8870\) \newline \tiny{(\(0.8309\)-\(0.9296\))}  &
	\(0.8908\) \newline \tiny{(\(0.8676\)-\(0.9136\))}  &
	\(0.8814\) \newline \tiny{(\(0.8179\)-\(0.9204\))}  &
	\(0.9003\) \newline \tiny{(\(0.8764\)-\(0.9209\))}  \\

	\midrule
	Cavity &
	\(0.9471\) \newline \tiny{(\(0.8705\)-\(1.0000\))}  &
	\(0.9375\) \newline \tiny{(\(0.6165\)-\(0.9845\))}  &
	\(0.9726\) \newline \tiny{(\(0.9588\)-\(0.9827\))}  &
	\(0.8750\) \newline \tiny{(\(0.6165\)-\(0.9845\))}  &
	\(0.9726\) \newline \tiny{(\(0.9588\)-\(0.9827\))}  \\

	\midrule
	Consolidation &
	\(0.9501\) \newline \tiny{(\(0.9208\)-\(0.9794\))}  &
	\(0.8857\) \newline \tiny{(\(0.8089\)-\(0.9395\))}  &
	\(0.8456\) \newline \tiny{(\(0.8198\)-\(0.8713\))}  &
	\(0.8476\) \newline \tiny{(\(0.7535\)-\(0.9028\))}  &
	\(0.9013\) \newline \tiny{(\(0.8783\)-\(0.9212\))}  \\

	\midrule
	Fibrosis &
	\(0.9300\) \newline \tiny{(\(0.8966\)-\(0.9634\))}  &
	\(0.9000\) \newline \tiny{(\(0.8281\)-\(0.9490\))}  &
	\(0.7557\) \newline \tiny{(\(0.7255\)-\(0.7865\))}  &
	\(0.8455\) \newline \tiny{(\(0.7538\)-\(0.9000\))}  &
	\(0.9025\) \newline \tiny{(\(0.8797\)-\(0.9223\))}  \\

	\midrule
	Hilar Enlargement &
	\(0.8854\) \newline \tiny{(\(0.8314\)-\(0.9393\))}  &
	\(0.8906\) \newline \tiny{(\(0.7875\)-\(0.9549\))}  &
	\(0.7246\) \newline \tiny{(\(0.6933\)-\(0.7568\))}  &
	\(0.6094\) \newline \tiny{(\(0.4637\)-\(0.7149\))}  &
	\(0.9061\) \newline \tiny{(\(0.8835\)-\(0.9255\))}  \\

	\midrule
	Nodule &
	\(0.9135\) \newline \tiny{(\(0.8686\)-\(0.9584\))}  &
	\(0.8630\) \newline \tiny{(\(0.7464\)-\(0.9223\))}  &
	\(0.9034\) \newline \tiny{(\(0.8807\)-\(0.9230\))}  &
	\(0.8630\) \newline \tiny{(\(0.7464\)-\(0.9223\))}  &
	\(0.9034\) \newline \tiny{(\(0.8807\)-\(0.9230\))}  \\

	\midrule
	Opacity &
	\(0.9412\) \newline \tiny{(\(0.9254\)-\(0.9570\))}  &	
	\(0.8966\) \newline \tiny{(\(0.8645\)-\(0.9233\))}  &
	\(0.8026\) \newline \tiny{(\(0.7726\)-\(0.8328\))}  &
	\(0.8404\) \newline \tiny{(\(0.8006\)-\(0.8712\))}  &
	\(0.9027\) \newline \tiny{(\(0.8783\)-\(0.9237\))}  \\

	\midrule
	Pleural Effusion &
	\(0.9805\) \newline \tiny{(\(0.9652\)-\(0.9957\))}  &
	\(0.9494\) \newline \tiny{(\(0.8946\)-\(0.9736\))}  &
	\(0.9035\) \newline \tiny{(\(0.8804\)-\(0.9235\))}  &
	\(0.8987\) \newline \tiny{(\(0.8408\)-\(0.9410\))}  &
	\(0.9635\) \newline \tiny{(\(0.9492\)-\(0.9767\))}  \\
	
	\bottomrule
	\end{tabularx}
	\smallskip
	\caption{Performance of the algorithms on CQ2000 dataset.}
	\label{table:results}
\end{table}

\begin{table}
	\centering
	\begin{tabularx}{\linewidth}{p{8.5em} p{4.9em} p{4.9em} p{4.9em} p{4.9em}p{4.9em}}
	\toprule

	\textbf{Finding} & \textbf{AUC \newline \small ($95\%$ CI)} &
	\multicolumn{2}{p{10em}}{\textbf{High sensitivity operating point}} &
	\multicolumn{2}{p{10em}}{\textbf{High specificity operating point}} \\

	& &
	Sensitivity&
	Specificity&
	Sensitivity&
	Specificity \\

	\midrule
	Normal(No Abnormality Detected) &
	\(0.8558\) \newline \tiny{(\(0.8531\)-\(0.8585\))}  &
	\(0.9010\) \newline \tiny{(\(0.8978\)-\(0.9041\))}  &
	\(0.5554\) \newline \tiny{(\(0.5516\)-\(0.5592\))}  &
	\(0.6561\) \newline \tiny{(\(0.6510\)-\(0.6611\))}  &
	\(0.9000\) \newline \tiny{(\(0.8977\)-\(0.9023\))}  \\

	\midrule
	Blunted CP angle &
	\(0.9471\) \newline \tiny{(\(0.9415\)-\(0.9528\))}  &
	\(0.9010\) \newline \tiny{(\(0.8896\)-\(0.9115\))}  &
	\(0.8817\) \newline \tiny{(\(0.8797\)-\(0.8837\))}  &
	\(0.8753\) \newline \tiny{(\(0.8628\)-\(0.8871\))}  &
	\(0.9001\) \newline \tiny{(\(0.8982\)-\(0.9020\))}  \\

	\midrule
	Cardiomegaly &
	\(0.9504\) \newline \tiny{(\(0.9465\)-\(0.9543\))}  &
	\(0.9010\) \newline \tiny{(\(0.8930\)-\(0.9085\))}  &
	\(0.8744\) \newline \tiny{(\(0.8723\)-\(0.8765\))}  &
	\(0.8494\) \newline \tiny{(\(0.8398\)-\(0.8583\))}  &
	\(0.9002\) \newline \tiny{(\(0.8982\)-\(0.9021\))}  \\

	\midrule
	Cavity &
	\(0.9642\) \newline \tiny{(\(0.9478\)-\(0.9805\))}  &
	\(0.9262\) \newline \tiny{(\(0.8859\)-\(0.9557\))}  &
	\(0.9003\) \newline \tiny{(\(0.8985\)-\(0.9022\))}  &
	\(0.8975\) \newline \tiny{(\(0.8525\)-\(0.9326\))}  &
	\(0.9395\) \newline \tiny{(\(0.9380\)-\(0.9410\))}  \\

	\midrule
	Consolidation &
	\(0.9414\) \newline \tiny{(\(0.9341\)-\(0.9486\))}  &
	\(0.9008\) \newline \tiny{(\(0.8867\)-\(0.9137\))}  &
	\(0.8774\) \newline \tiny{(\(0.8754\)-\(0.8795\))}  &
	\(0.8732\) \newline \tiny{(\(0.8571\)-\(0.8872\))}  &
	\(0.9002\) \newline \tiny{(\(0.8983\)-\(0.9021\))}  \\

	\midrule
	Fibrosis &
	\(0.9368\) \newline \tiny{(\(0.9277\)-\(0.9460\))}  &
	\(0.9009\) \newline \tiny{(\(0.8835\)-\(0.9166\))}  &
	\(0.8280\) \newline \tiny{(\(0.8256\)-\(0.8303\))}  &
	\(0.8453\) \newline \tiny{(\(0.8246\)-\(0.8644\))}  &
	\(0.9000\) \newline \tiny{(\(0.8981\)-\(0.9019\))}  \\

	\midrule
	Hilar Enlargement &
	\(0.8442\) \newline \tiny{(\(0.8282\)-\(0.8602\))}  &
	\(0.9006\) \newline \tiny{(\(0.8792\)-\(0.9193\))}  &
	\(0.5649\) \newline \tiny{(\(0.5618\)-\(0.5680\))}  &
	\(0.5790\) \newline \tiny{(\(0.5450\)-\(0.6103\))}  &
	\(0.9006\) \newline \tiny{(\(0.8987\)-\(0.9024\))}  \\

	\midrule
	Nodule &
	\(0.9202\) \newline \tiny{(\(0.9094\)-\(0.9310\))}  &
	\(0.9003\) \newline \tiny{(\(0.8817\)-\(0.9170\))}  &
	\(0.7735\) \newline \tiny{(\(0.7709\)-\(0.7762\))}  &
	\(0.7964\) \newline \tiny{(\(0.7721\)-\(0.8192\))}  &
	\(0.9005\) \newline \tiny{(\(0.8987\)-\(0.9024\))}  \\

	\midrule
	Opacity &
	\(0.9357\) \newline \tiny{(\(0.9326\)-\(0.9388\))}  &	
	\(0.9010\) \newline \tiny{(\(0.8954\)-\(0.9063\))}  &
	\(0.8215\) \newline \tiny{(\(0.8189\)-\(0.8240\))}  &
	\(0.8412\) \newline \tiny{(\(0.8343\)-\(0.8477\))}  &
	\(0.9000\) \newline \tiny{(\(0.8980\)-\(0.9020\))}  \\

	\midrule
	Pleural Effusion &
	\(0.9566\) \newline \tiny{(\(0.9527\)-\(0.9605\))}  &
	\(0.9006\) \newline \tiny{(\(0.8921\)-\(0.9086\))}  &
	\(0.8776\) \newline \tiny{(\(0.8755\)-\(0.8797\))}  &
	\(0.8755\) \newline \tiny{(\(0.8660\)-\(0.8842\))}  &
	\(0.9000\) \newline \tiny{(\(0.8981\)-\(0.9019\))}  \\
	
	\bottomrule
	\end{tabularx}
	\smallskip
	\caption{Performance of the algorithms on CQ100k dataset}
	\label{table:results90k}
\end{table}

\section{Discussion}
Long before deep learning, automated chest X-ray interpretation using traditional image processing methods have been used to identify chest-ray views, segment parts of the lung, identify cardiomegaly or lung nodules and diagnose tuberculosis. However, these traditional methods did not come into routine clinical use because of their need for standardized X-ray quality, machine model and images free of artefacts\cite{xue2015chest,duryea1995fully,yousefian2015automated,xu1997development,jaeger2014automatic,hogeweg2015automatic}. The success of deep learning for image interpretation renewed the interest in automated chest X-ray interpretation with many research groups leveraging convolutional neural networks to detect pneumonia, tuberculosis and other chest diseases\cite{shin2016learning,lakhani2017deep,rajpurkar2017chexnet,wang2017chestx,li2017thoracic,bar2015deep,hwang2019development}. 

In 2017, \citet{shin2016learning,lakhani2017deep} bought attention to the use of deep learning to interpret chest X-ray images. \citet{wang2017chestx} used the ChestX-ray14 dataset, a single-source dataset containing 112,120 X-rays and NLP-generated labels for 14 thoracic diseases that was made publicly available by the NIH. Several groups used this dataset to train and validate deep learning algorithms with NLP-generated labels as ground truth and reported AUCs ranging from 0.69 to 0.91 for various abnormalities\cite{rajpurkar2017chexnet,wang2017chestx,li2017thoracic,bar2015deep}. Using the same ChestX-ray14 dataset, \citet{rajpurkar2017chexnet} trained their algorithm `CheXNet' to detect pneumonia and validated it against a dataset of 420 X-rays independently reported by Stanford Radiologists. They found that the algorithm outperformed Radiologists in detecting pneumonia\cite{rajpurkar2017chexnet}. More recently, 3 large public single-source datasets from about 400,000 patients/imaging studies along with NLP extracted labels were released\cite{irvin2019chexpert,johnson2019mimic,bustos2019padchest}. These are decently sized but contain a lot of follow-up scans from the same patient reducing the effective variability. Additionally, a significant portion of these datasets contain artifacts due to being in-patient follow up scans limiting their usefulness for training robust models. However, these datasets will inspire a lot of research particularly by combining domain specific knowledge and clinical insights.

To the best of our knowledge, ours is the largest chest X-ray training and testing dataset reported in the literature: we trained a deep learning algorithm on 2.3 million chest X-rays, and validated it on two datasets, one with 2000 chest X-rays where the gold standard is the majority of 3 radiologist reads and another that has 100,000 X-rays where the gold standard is the radiologist report.  The training dataset was sourced from a large number of centres. Since a Chest X-ray is not standardized, this ensures that most variability in terms of manufacturers and settings is captured potentially making the algorithms more robust in new scenarios. On the CQ2000 dataset, The algorithm achieved an AUC of 0.92 for differentiating normal from abnormal chest X-rays, while the AUC for detection of individual abnormalities varied from 0.89 to 0.98. AUCs were higher for abnormalities with higher prevalence in the dataset, greater inter-reader agreement and consistency of reporting terminology. The highest accuracy was achieved for findings that were unambiguously defined and reported with consistent terminology, such as pleural effusion or cardiomegaly. Unlike previous algorithms trained to make a diagnosis, our chest X-ray algorithm was trained to identify abnormal findings and differentiate normal from abnormal. We did this to facilitate clinical use of the algorithm across geographies, independent of local disease prevalence. The AUCs are similar in both the CQ2000 and CQ100k datasets for all findings except for detecting normals where the algorithm performed much better on the CQ2000 dataset. The sensitivity at the same specificity was significantly lower on the CQ100k dataset. We expect that this is because a lot of insignificant/subtle findings like sutures, aortic arch calcifications or increased vascularity which are treated as abnormal by the NLP labeler may not have been reported by the raters of the CQ2000 dataset. The CQ2000 dataset was also enriched with abnormalities that are focussed on in this study leading to a higher AUC. The similar AUCs on both datasets across the target findings indicate that the NLP labeler was fairly accurate and that the enrichment done for statistical reasons was not to the algorithms advantage.

There are several limitations to this study. The CQ100k dataset had sufficient positive samples for all target findings. Although CQ2000 dataset has been enriched with the strategy described in section \ref{sec:datasets} which allowed substantial positive samples for most target findings, there were not enough positive cases to reliably report the accuracy of cavity which resulted in wide confidence intervals for sensitivity. An NLP algorithm was used for enrichment in CQ2000 which might have induced a selection bias. This is because the same NLP algorithm was used to both provide the training labels for the algorithm as well as enrich the validation set. However, this risk is minimal as the accuracy of the NLP algorithm was very high when validated by expert readers. In the CQ2000 dataset, multiple radiologist reads were used to establish the gold standard. The reliability of the gold standard established in such a manner is, qualitatively, a function of the concordance of the raters. Although reader consensus is the customary mode of establishing ground truth in radiology, it does not provide information on inter-reader variability\cite{bankier2010consensus,obuchowski1996simple}. We used a 3-reader majority opinion, without consensus, as ground truth. The concordance, sensitivity and specificity of chest X-ray interpretations by radiologists are known to vary widely depending on the abnormality being detected, reader expertise, and clinical setting\cite{moncada2011reading,hopstaken2004inter,ochsmann2010inter,zellweger2006intra,kosack2017evaluation,swingler2005diagnostic,bourbeau1988between}. Though, the inter-reader variability we encountered for the CQ2000 dataset for various abnormalities is similar to that previously documented, these are not very high. Agreement rates for opacity are also low in our study, likely related to differences in readers interpretation of our definition of `opacity'. Additionally, Our readers were blinded to clinical history, a factor that might have impacted both accuracy and inter-reader variance. 

Experience did not reduce inter-reader variance, suggesting that the high inter-reader variance is due to the inherent difficulty in interpreting a 2D greyscale display of a 3D complex body part containing tissue densities that range from air to bone. Another limitation in this study is that we did not exclude multiple X-rays from the same patient; however on the CQ2000 dataset, given that bedside and portable X-rays were excluded, the probability that repeated X-rays from the same patient occurred in this dataset is very low whereas on the CQ100k dataset, the ratio of the number of studies to patients is very low($10:9$). This study is also not comprehensive. It does not include validation of many significant abnormal findings that are detectable on a chest X-ray most notably pneumothorax and rib fractures because of the research groups' focus on findings that can be used to screen infections. Though the algorithms produce heatmaps and bounding boxes, a missing component in this study is the validation of this ability to localize the lesions. This is a vital component of a potential clinical decision support system.

Radiologist opinion is not the ground truth for chest X-rays but is only a precursor to the final clinical and histopathology diagnosis. We tested the algorithm to determine if it can replicate radiologist observation of abnormality on chest X-rays rather than diagnose chest pathology. Our study demonstrates that deep learning algorithms trained on large datasets with manually curated, fully automated NLP based labels can accurately detect abnormalities on chest X-rays, with close to radiologist-level accuracy. Apart from providing automated chest X-ray interpretations in under-served and remote locations, automated draft or preliminary reports of abnormal chest X-rays can improve turnaround time for reporting and clear backlogs. Further research is required to assess the clinical acceptance of artificial intelligence in the real-world and quantify the benefit of such automation to physicians and their patients.

\bibliographystyle{unsrtnat}
\bibliography{main}

\end{document}